\documentclass{article} 
\usepackage{iclr2017_workshop,times}
\usepackage{hyperref}
\usepackage{url}
\usepackage{graphicx}

\title{Train Once, Test Anywhere: Zero-Shot Learning for Text Classification}

\author{Pushpankar Kumar Pushp \& Muktabh Mayank Srivastava \\
ParallelDots, Inc.\thanks{www.paralleldots.xyz} \\
\texttt{\{pushpankar,muktabh\}@paralleldots.com} \\
}

\begin{document}

\maketitle

\begin{abstract}
Zero-shot Learners are models capable of predicting unseen classes. In this work, we propose a Zero-shot Learning approach for text categorization. Our method involves training model on a large corpus of sentences to learn the relationship between a sentence and embedding of sentence's tags. Learning such relationship makes the model generalize to unseen sentences, tags, and even new datasets provided they can be put into same embedding space. The model learns to predict whether a given sentence is related to a tag or not; unlike other classifiers that learn to classify the sentence as one of the possible classes. We propose three different neural networks for the task and report their accuracy on the test set of the dataset used for training them as well as two other standard datasets for which no retraining was done. We show that our models generalize well across new unseen classes in both cases. Although the models do not achieve the accuracy level of the state of the art supervised models, yet it evidently is a step forward towards general intelligence in natural language processing.
\end{abstract}

\section{Introduction}
\begin{figure}[h]
  \centering
  \fbox{
  	\rule[-.5cm]{0cm}{4cm} 
  		\includegraphics[width=8cm]{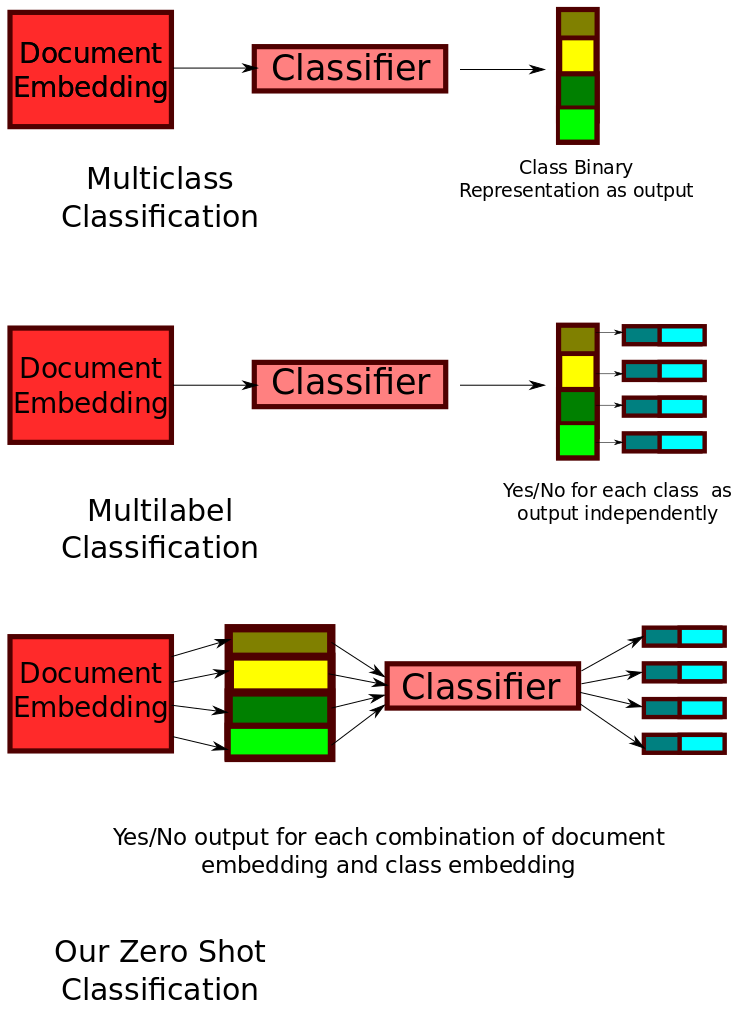} 
    \rule[-.5cm]{0cm}{0cm}}
  \label{technique}
  \caption{Difference between multiclass, multilabel classification and zero-shot classification framework proposed}
\end{figure}

Zero-shot learning has been an area of special interest in recent years. It not only allows scaling of algorithms across unseen classes but also can be used across datasets as we try to show in this work. In this work, we report a methodology that can be used for zero-shot learning in the case of text categorization. To achieve this, we model the task of text categorization as a binary classification problem of finding relatedness between sentences and categories. Models trained in this fashion learn the relatedness (yes/no) of the given sentence for each category separately, instead of predicting the given class as in multiclass-multilabel classification (Figure \ref{technique}). For instance, the sentence - “Obama said that GOP’s efforts to repeal health care were aggravating” is trained to belong to categories ‘politics’ and ‘healthcare’ but not to categories ‘sports’ and ‘technology’. Table 1 lists other examples of sentences belonging to multiple and overlapping text categories. The proposed redefinition allows neural network model trained on one dataset to be deployed for text categorization on other datasets without the need of retraining the model.

We further propose neural network architectures that can be used as zero-shot learners with this technique. While one model is a single layered neural network on mean of word embeddings, the other two models use LSTMs to model sentence as a sequence.  

We believe that training networks with a large amount of data with noisy annotation leads to more generalized models as compared to training with smaller datasets that are annotated specifically for underlying tasks. Moreover, utilization of noisy annotated data from open web saves annotation cost. Therefore, we trained our model with news headlines crawled from around the web with their Search Engine Optimization (SEO) tags as categories (also called the source dataset) and test its performance. We further test our model on News aggregator \cite{Lichman2013UCIRepository} and tweet classification \cite{TweetClassification} datasets, hence showing the concept of relatedness it learns is useful across datasets. Briefly, the contributions of this paper are three-fold:-

\begin{enumerate}
\item We propose a zero-shot learning framework for zero shot text categorization as binary classification task to find relatedness between text and categories. We show that this framework can adapt to any number of text categories as well as across datasets, without the need of re-training or fine-tuning the model.
\item We propose three neural network architectures that can use the technique above and can be used for zero-shot classification.
\item We report accuracy of the zero-shot learning capability of our model trained on source dataset on different datasets and compare it with state-of-the-art results obtained through models that were specifically trained on those datasets. We show that our architecture can generalize to classes it has not seen and even datasets it has not been trained on.
\end{enumerate}

\section{Related Work}
\label{gen_inst}
Many zero-shot learning approaches have been proposed in the domain of computer vision \cite{Sandouk2016Multi-LabelEmbedding}, \cite{Socher2013Zero-ShotTransfer}. However, there exists a very limited amount of work towards zero-shot learning in the domain of natural language processing. To the best of our understanding, this is the first work to report a zero shot learning solution for text categorization.
Our first architecture is a single layered neural network on concatenation of 1. mean embedding of the sentence and 2. the embedding of the tag. It is inspired by shallow architectures which get good scores on text classification tasks like \cite{joulin2016fasttext}. The second architecture, instead of taking a mean of embeddings before passing it to classification layer, tries to model the sequence using an LSTM \cite{Hochreiter1997LONGMEMORY} .
Our third LSTM architecture may be considered similar to the architecture used by a \cite{Wang2016Attention-basedClassification} for aspect-based sentiment analysis. Instead of the "Aspect Term", we pass the embedding of the tag to be considered related/not-related. However, we do not employ the component of attention as mentioned in the work by Wang et. al.

\section{Data}
\label{headings}
\subsection{Source Dataset}
We crawled more than 4,200,000 news headlines from around the web along with their SEO tags. The corpus had more than 300,000 unique SEO tags. For simplicity, we henceforth refer to news headlines and SEO tags as sentences and tags, respectively. A news article can have more than one SEO tags. In such cases, we added multiple instances of the sentence to our data, one for each tag. 

During data preparation, we fixed the sentence length to contain 28 words by truncating longer sentences and repeating words for shorter sentences. The Source Dataset was split into a train and test set.

\begin{table}
	\caption{Sample Train data}
    \label{Sample Train data}
    \centering
    \begin{tabular}{p{10cm}p{3cm}}
        \multicolumn{1}{c}{\bf Sentence}  &\multicolumn{1}{c}{\bf Tags}\\
        Mastercard on Tuesday said it is extending its relationship with online payments processor PayPal as part of a multi-faceted deal that'll be mutually beneficial to all parties involved. & mastercard, paypal, online payment, merchant\\
        \\
        THE internet is blowing up all over again - and it's all the fault of a pair of thongs. & editor's pick, humor, internet, Read More\\
        \\
        LUKE Aglio believes he can make you feel stronger, more flexible and rid you of pain and tension without having to lift a weight or spending years doing yoga. & muscle activation, health, alto performance, luke aglio\\
    \end{tabular}
\end{table}

\subsection{Test Datasets}
 The algorithm was trained on the training set of the source dataset and its accuracy of text categorization was tested on two other open datasets. We used UCI News Aggregator and tweet classification datasets as test datasets.
\subsubsection{UCI News Aggregator}
The dataset contained more than 420,000 sentences belonging to four categories: technology, business, medicine, and entertainment. We report our accuracy on the entire dataset. Since the granularity level of categories is different from SEO tags in Source Dataset, we use the concept of category tree for UCI-Aggregator dataset (Refer Testing section for more details).
\subsubsection{Tweet Classification}
The dataset contains 6 categories: business, health, politics, sports, technology, and entertainment. The dataset has 1993 sentences and we used all the sentence for testing. Our model can classify the dataset directly using embeddings of categories (health, politics etc.), but we get even better results using category tree as for UCI-Aggregator dataset.
\section{Architectures}
Three different architectures are tried out for zero-shot learning and results are reported. They are referred to as Architecture 1, 2, and 3 respectively. We initialized word embedding with a pretrained embedding \cite{GoogleHosting.} for all three. For notation, let’s consider the tag's embedding is $T_E$ and the word embeddings of the sentence are [$S_1$, $S_2$ .. $S_N$].
\subsection{Architecture 1}
 We concatenate the dimensionwise mean of [$S_1$, $S_2$ .. $S_N$] with $T_E$ and pass it thorugh a fully connected layer to classify if sentence and tag are related.
\begin{figure}[h]
  \centering
  \fbox{
  	\rule[-.5cm]{0cm}{4cm} 
  		\includegraphics[width=8cm]{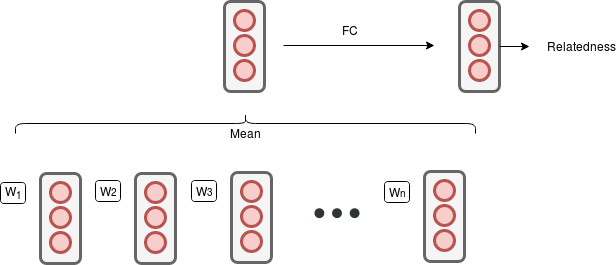} 
    \rule[-.5cm]{0cm}{0cm}}
  
  \caption{Architecture 1}
\end{figure}
\subsection{Architecture 2}
The input sent at a time step t, to the LSTM is [$S_t$], where $S_t$ is the embedding of $t^{th}$ word of the sentence. We concatenate the last hidden state of the network with $T_E$ and pass it thorough a fully connected layer to classify if sentence and tag are related.
\begin{figure}[h]
  \centering
  \fbox{
  	\rule[-.5cm]{0cm}{4cm} 
  		\includegraphics[width=8cm]{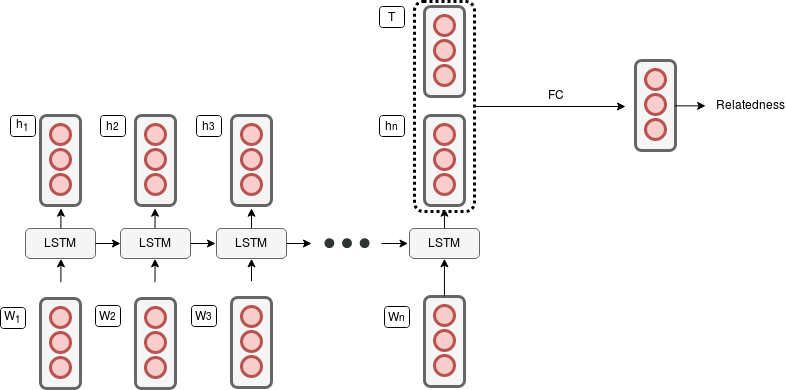} 
    \rule[-.5cm]{0cm}{0cm}}
  
  \caption{Architecture 2}
\end{figure}
\subsection{Architecture 3}
 The input sent at a time step t, to the LSTM is [$T_E$:$S_t$], where $S_t$ is the embedding of $t^{th}$ word of the sentence. We use the last hidden layer of LSTM and predict if it is related to tag’s embedding $T_E$.

\begin{figure}[h]
  \centering
  \fbox{
  	\rule[-.5cm]{0cm}{4cm} 
  		\includegraphics[width=8cm]{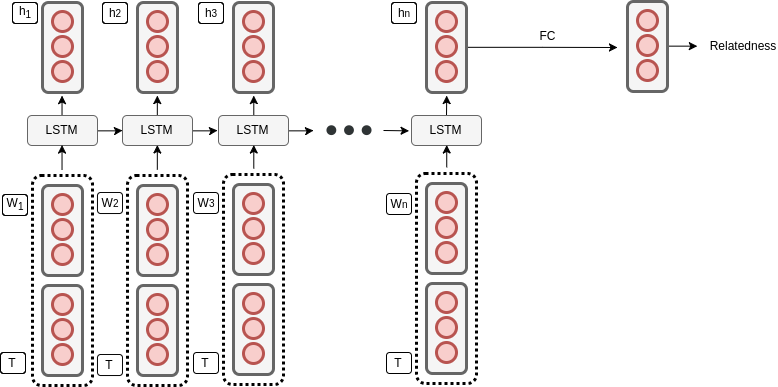} 
    \rule[-.5cm]{0cm}{0cm}}
  
  \caption{Architecture 3}
\end{figure}

\section{Training}
Our dataset contained more than 300,000 tags, making the approach of training with multi-class classification intractable. Therefore, we converted the tag prediction task to binary classification task, where the model predicts whether a given sentence is related to given tag or not. In other words, we train our algorithm for generic knowledge of whether a given sentence is related to a particular class or not. This is different from  the knowledge "which class among a set of classes does a particular sentence belongs to?", learned by a normal multi-class classifier.  Figure \ref{technique} tries to explain this visually. We trained each sentence with 50\% related and 50\% randomly selected unrelated tags. We trained the model for binary cross entropy loss with Adam optimizer \cite{Kingma2014Adam:Optimization}.

\section{Testing}
We tested our models on UCI News Aggregator and Tweet Classification dataset.
There is a slight difference between text classes used in these datasets and SEO tags of our Source Dataset, SEO tags are more atomic concepts as compared to UCI classes. For example, while SEO tags for a sentence "Bitcoin futures could open the floodgates for institutional investors" would be Bitcoin, Commodity, Futures, Cryptocurrency, Hedge Funds, and Mutual Funds, the tags in UCI/Tweet Categorizer dataset would be classes representing broader concepts i.e. Business and Technology. Therefore, to test the accuracy of our classifier on these classes, we first create a category tree for these datasets to list multiple tags that would belong to each class. For example, tags like "forex", "financial markets", "stocks", "production" and "Business" itself might belong to category tree under "Business" class. To predict relatedness of a given sentence to a particular class, we first predict the relatedness probability of all the tags listed for that class and take their mean. The sentence is then classified into the classes that have mean relatedness probability above a certain threshold for that sentence. The threshold of relatedness score is a hyperparameter. This technique allowed our model to function across different levels of granularity of the concepts in which text can be classified. Making the category tree for either datasets is just a work of few minutes, thinking of what all concepts might belong to a particular class and listing them down. For example, when testing on Tweet Categorization dataset, the category tree is just three tags per class. 

\section{Results}
\label{others}
The models trained using architectures 1, 2 and 3 achieved  \textbf{72\%}, \textbf{72.6\%} and \textbf{74\%} accuracy respectively on test set of Source Dataset for the binary classification task. For the tags which are not present in the training set and only in the test set, the accuracy is even higher at \textbf{78\%},\textbf{76\%} and \textbf{81\%} respectively. Further, the models achieve \textbf{61.73\%}, \textbf{63\%} and \textbf{64.21\%} accuracy respectively on the News Aggregator dataset using a category tree at the threshold of 0.5 on relatedness score. The reported accuracy is much lesser than the state-of-the-art accuracy (94.75\%) \cite{ClassifyingKaggle} on this dataset. However, considering that our model had not even seen a single sample from the given dataset as opposed to fully supervised methods, the reported results are still remarkable.

We evaluated the performance of our model on tweet classification dataset using a category tree on a threshold of 0.5 relatedness scores. Architecture 1,2 and 3 got  \textbf{64\%}, \textbf{53\%} and \textbf{64.5\%} accuracy on the dataset. In contrast to best results of supervised model like SVC and multinomial naive bayes, which have 74\% and 78\% accuracies respectively,our models are not trained on the dataset. If we do not use a category tree and use direct class names to classify, we can still get \textbf{49\%} accuracy using architecture 3.

\begin{table}
	\caption{Category Tree for News Aggregator dataset}
    \label{Category Tree label}
    \centering
    
	\begin{tabular}{|l|l|l|l|l|}
        \textbf{Catgeory Names} & business & technology & entertainment & medicine\\
        \hline
        
        \textbf{Words for the category} & finance & internet & actor & health\\
        & revenue & android & music & disease \\
        
	\end{tabular}
\end{table}

\section{Conclusion}	

In this work, 
we introduce techniques and models that can be used for zero-shot classification in text. We show that our models can get better than random classification accuracies on datasets without seeing even one example. We can say that this technique learns the concept of relatedness between a sentence and a word that can be extended beyond datasets. That said, the levels of accuracy leave a lot of scope for future work.

\bibliography{iclr2017_workshop}
\bibliographystyle{iclr2017_workshop}

\end{document}


\begin{table}
	\caption{Category Tree for Tweet classification}
    \label{Category Tree label}
    \centering
    
	\begin{tabular}{|l|l|l|l|l|l|l|}
        \textbf{Catgeory Names} & health & sports & entertainment & business & technology & politics\\
        \hline
        
        \textbf{Words for the category} & health & sports & entertainment & business & technology & politics\\
        & medicine & game & movie & money & internet  & election\\
        & doctor & football & actor & investment & computer & government\\
        
	\end{tabular}
\end{table}